\documentclass[preprint,12pt]{elsarticle}
\usepackage{amssymb}
\usepackage{amsmath}
\usepackage{algorithmic}
\usepackage[ruled]{algorithm2e} % For algorithms
\usepackage{graphicx}
\usepackage{textcomp}
\usepackage{makecell}
\usepackage{multicol}
\usepackage{multirow}

\journal{arXiv}
\begin{document}
\begin{frontmatter}

\title{
Vortex Feature Positioning: \\Bridging Tabular IIoT Data and Image-Based Deep Learning
%Converting Tabular Data into Images with Attribute Correlation for CNNs
}
%\author{Anonymous Author(s)}
\author[cmu]{Jong-Ik Park}
\author[cau]{Sihoon Seong}
\author[essex]{JunKyu Lee}
\author[cau]{Cheol-Ho Hong\corref{cor1}}

\cortext[cor1]{Corresponding author}

\affiliation[cmu]{organization={Department of Electrical and Computer Engineering, Carnegie Mellon University}, %Department and Organization
            addressline={\\5000 Forbes Ave}, 
            city={Pittsburgh},
            postcode={15213}, 
            state={PA},
            country={United States}}

\affiliation[cau]{organization={Department of Intelligent Semiconductor Engineering, Chung-Ang University}, %Department and Organization
            addressline={\\84 Heukseok-ro, Dongjak District}, 
            city={Seoul},
            postcode={06974}, 
            country={Korea}}

\affiliation[essex]{organization={Institute for Analytics and Data Science, University of Essex}, %Department and Organization
            addressline={\\Wivenhoe Park}, 
            city={Colchester},
            postcode={CO4 3SQ}, 
            country={United Kingdom}}

\begin{abstract}
Tabular data from IIoT devices are typically analyzed using decision tree-based machine learning techniques, which struggle with high-dimensional and numeric data. To overcome these limitations, techniques converting tabular data into images have been developed, leveraging the strengths of image-based deep learning approaches such as Convolutional Neural Networks. These methods cluster similar features into distinct image areas with fixed sizes, regardless of the number of features, resembling actual photographs. However, this increases the possibility of overfitting, as similar features, when selected carefully in a tabular format, are often discarded to prevent this issue. Additionally, fixed image sizes can lead to wasted pixels with fewer features, resulting in computational inefficiency. We introduce Vortex Feature Positioning (VFP) to address these issues. VFP arranges features based on their correlation, spacing similar ones in a vortex pattern from the image center, with the image size determined by the attribute count. VFP outperforms traditional machine learning methods and existing conversion techniques in tests across seven datasets with varying real-valued attributes.
\end{abstract}


\end{frontmatter}

\section{Introduction} \label{introduction}
The industrial Internet of Things (IIoT) collects vast amounts of sensor data commonly presented in a tabular form~\cite{wang2018industrial}. This industry data is often analyzed using traditional machine learning (ML) techniques based on decision tree algorithms~\cite{quinlan1986induction}, supported by voting algorithms~\cite{dietterich2000ensemble}, and enhanced by ensemble techniques such as Gradient Boosting~\cite{friedman2002stochastic},  XGBoost~\cite{chen2016xgboost}, LightGBM~\cite{ke2017lightgbm}, and CatBoost~\cite{prokhorenkova2018catboost}. These techniques identify defective goods and detect anomalies in IIoT systems. However, the increasing complexity and breadth of attributes in IIoT data, along with the need for high-resolution data, often exceed the capabilities of traditional ML methods. For example, in predictive maintenance, sensors may record high-frequency vibration data that is highly granular, posing a challenge for traditional ML methods to process~\cite{borisov2022deep}. Similarly, in optimizing wafer fabrication processes, precise control of numerous parameters based on real-time sensor data is crucial for yield improvement, requiring more sophisticated solutions~\cite{kitchat2021defective}. 

When the input data is numeric and highly complex, a deep learning approach becomes more suitable for analyzing tabular data~\cite{babaev2019et-rnn}. Many researchers have explored using Convolutional Neural Networks (CNNs)~\cite{borisov2021deep} by converting tabular data into images. CNNs are a deep learning technique with numerous parameters, excelling particularly in handling image data that exhibit distinct patterns within pixels~\cite{zeiler2014visualizing}. However, applying CNNs to tabular data is challenging because convolution kernels are designed to detect specific spatial patterns in multidimensional data. Typically, tabular data is represented as a vector, and even when reshaped into a 2-D matrix using conventional methods, the resulting `image' often lacks meaningful spatial patterns. Consequently, for CNNs to perform effectively, the attributes of the tabular data must be strategically arranged in the reshaped matrix to create recognizable spatial patterns~\cite{borisov2021deep, grinsztajn2022tree}.

Previous studies, such as DeepInsight~\cite{sharma2019deepinsight}, REFINED~\cite{bazgir2020representation}, IGTD~\cite{zhu2021converting}, and SuperTML~\cite{sun2019supertml}, have proposed methods to convert 1-D tabular data into 2-D image data by considering the positions of attributes. Specifically, DeepInsight, REFINED, and IGTD focus on grouping similar attributes with high correlations in specific locations within a 2-D matrix. This enables a CNN model to learn patterns within these similar attributes. However, this approach can lead to overfitting if similar attributes are intensively grouped in one area, making it difficult for the model to train on universal patterns between dissimilar attributes~\cite{moon2020improved}. Additionally, these methods convert tabular data into fixed-size images, which can result in wasted pixels for smaller datasets or insufficient space to represent all attribute values for larger datasets. Wasted pixels lead to computational inefficiency, as the CNN processes empty or irrelevant areas of the image, consuming resources without contributing to learning valuable patterns. Conversely, SuperTML carves tabular features (or attributes) of a sample (or row) onto an empty black image by assigning different font sizes to features based on their importance. While SuperTML has shown improved performance on some datasets, it has limitations: it may not be applicable when there are too many attributes to fit on a reasonably sized image, and its performance can vary depending on the font type.

This paper presents a novel approach, Vortex Feature Positioning (VFP), which converts tabular data into images tailored for CNNs while accounting for attribute correlations. With the increasing number of sensors in IIoT, there is a corresponding rise in attributes with real value sensor data~\cite{krishnamurthi2020overview}. To accommodate such IIoT applications, VFP arranges the attributes of the tabular data in a vortex shape, rotating from the center based on their Pearson correlation coefficient (PCC). This arrangement facilitates convolution operations, allowing for the optimal extraction of essential patterns. By positioning low-correlated features near the center of the converted image, VFP creates a convex-like loss landscape.
A convex-like loss landscape helps prevent overfitting by ensuring that the optimization process is smoother and less likely to get stuck in local minima. It encourages the model to find broader, more generalizable solutions rather than narrowly focusing on specific patterns that may not apply to unseen data. 
Additionally, this vortex shaping allows for the formation of images with flexible sizes, ensuring that the spatial representation is optimally adapted to the number of attributes. This flexibility further contributes to the efficient training of the CNN model.

\begin{table}[b]
    \centering 
    \caption{Overview of datasets in this study, detailing the number of attributes, sample count, and labels, and noting any missing data.}
    \label{table:data description}
    \includegraphics[width=\linewidth]{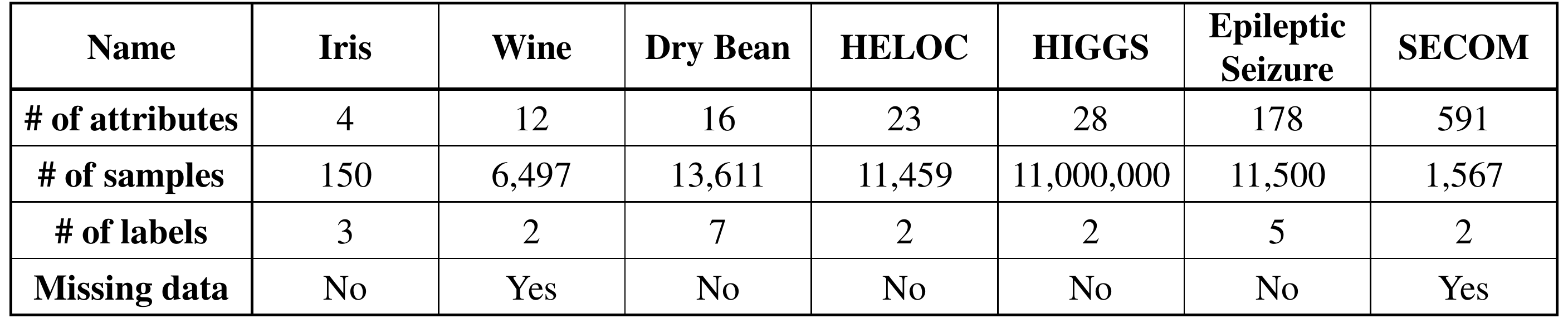}
\end{table}

We evaluated VFP on datasets related to typical industrial environments and general datasets to ensure its applicability across a broad range of IIoT environments. These datasets were sourced from the UCI Machine Learning Repository~\cite{uci_ml_repo} and Kaggle~\cite{kaggle}. The industrial datasets include Wine, Dry Bean, and SECOM, with SECOM specifically containing semiconductor manufacturing data. For general IIoT applications, we included Epileptic Seizure, Iris, HELOC, and HIGGS. Detailed descriptions of each dataset are provided in Table~\ref{table:data description}. As shown in the evaluation section, VFP outperforms traditional ML techniques such as XGBoost and CatBoost, and it generally excels across most datasets compared to techniques utilizing CNNs, such as DeepInsight, REFINED, and IGTD.

Machine and deep learning approaches aim to effectively discover complex but generalized patterns in training data. In this context, VFP makes three significant contributions:
\begin{itemize}
\item VFP outperforms traditional ML techniques for tabular data and previous methods for converting tabular data into images for CNNs.
\item VFP can convert tabular data with varying numbers of attributes into images with an optimized number of pixels.
\item Since VFP is a data format transformation technique, it can be utilized with any state-of-the-art CNNs and training methods.
\end{itemize}

With these contributions, VFP presents a valuable and versatile tool for transforming tabular data into images suitable for use with CNNs.

The rest of the paper is structured as follows: In Section~\ref{sec: related work}, we discuss related works. We present our proposed VFP method in Section~\ref{sec: voltex positioning}. Section~\ref{sec: convergence} shows the convergence analysis of VFP, Section~\ref{sec: evaluation} describes the experimental evaluation, and in Section~\ref{sec: conclusion}, we conclude the paper and discuss future work.

\section{Related Work} \label{sec: related work}
\subsection{Machine Learning Techniques for Tabular Data} \label{subsec: MLFT}
Traditional ML methods for analyzing tabular data include Gradient Boosting~\cite{friedman2002stochastic}, XGBoost~\cite{chen2016xgboost}, LightGBM~\cite{ke2017lightgbm}, and CatBoost~\cite{prokhorenkova2018catboost}, all based on decision tree algorithms~\cite{quinlan1986induction}. 
These techniques are widely used and remain dominant over CNNs when handling tabular data~\cite{shwartz2022tabular}. They mitigate overfitting in regression or classification tasks by employing ensemble methods such as boosting and bagging~\cite{friedman2002stochastic}. 
However, the performance of traditional ML techniques decreases as the number of attributes increases. The increased number of attributes negatively impacts testing performance, as the models struggle to generalize well on unseen data due to high dimensionality and complexity~\cite{prokhorenkova2018catboost}. Additionally, the training speed decreases when the attributes consist of numerous real values, as seen in industrial sensor data~\cite{wang2018industrial}. This decline is due to the exponential growth in the number of branches in the tree, leading to slow training speeds and high computational costs~\cite{ke2017lightgbm}. LightGBM was introduced with the leaf-wise algorithm to improve training speed. Still, it generally does not outperform XGBoost and CatBoost because it cannot capture the detailed feature information available in the level-wise algorithm~\cite{prokhorenkova2018catboost}. Therefore, finding a new method to handle tabular data effectively with numerous real attributes is necessary.

\subsection{Converting Tabular Data into Images for CNNs} \label{subsec: converting tabular into images}
Industries such as semiconductor manufacturing generate extensive tabular data characterized by real-number attributes, which differ from integer-based categorical attributes. As the number of attributes increases, this characteristic can hinder the performance of traditional ML techniques~\cite{wang2018industrial, borisov2021deep}.

CNNs can handle complex patterns in feature maps with real values, but their effectiveness is primarily limited to 2-D spatial data, such as images~\cite{zeiler2014visualizing}, rather than 1-D convolution operations~\cite{borisov2021deep, shwartz2022tabular}. Therefore, researchers have proposed various methods to convert tabular data into images for CNNs.
Methods such as DeepInsight~\cite{sharma2019deepinsight}, REFINED~\cite{bazgir2020representation}, IGTD~\cite{zhu2021converting}, and SuperTML~\cite{sun2019supertml} have demonstrated superior performance compared to traditional ML approaches, especially when there are many attributes in tabular data.
However, these methods may result in overfitting or the oversight of global patterns~\cite{borisov2021deep}. A key aspect of these methods is that each attribute is assigned a fixed position within a 2D (or 3D) matrix by grouping similar attributes. However, grouping similar attributes could hinder learning complex patterns across various attributes, as convolutional operations might overly focus on patterns of adjacent, similar features~\cite{dumoulin2016guide, borisov2021deep}. This strategy contrasts sharply with traditional ML practices that recommend dropping highly correlated attributes to prevent overfitting~\cite{yu2003feature, borisov2021deep}.

SuperTML~\cite{sun2019supertml} engraves features of tabular data onto an empty black image (i.e., a zero 2-D matrix), with each feature engraved in varying sizes based on its importance. Although SuperTML performed better than traditional ML techniques in some datasets, it faces challenges in determining the extent to increase image size with a growing number of attributes. It also requires prioritizing attributes due to considerations of font size when the choice of font type influences engraving features and its performance. These constraints highlight the necessity for a more generalized method.

Despite the enhanced performance achieved by DeepInsight, REFINED, IGTD, and SuperTML, it remains to be seen whether the improvement stems from CNNs' ability to detect complex patterns through 2-D convolution operations or primarily from their feature positioning methods. To address this uncertainty, we introduce eight different feature positioning scenarios in CNNs in Section~\ref{sec: voltex positioning}. These scenarios aim to demonstrate that the effectiveness of CNNs in detecting complex patterns is not solely responsible for improved test performance; how features are positioned also plays a critical role. Based on these insights, we introduce a new method for positioning features called Vortex Feature Positioning (VFP). VFP directly transforms all attributes of tabular data into images based on their correlations, reducing the risk of overfitting by considering the varying degrees of correlations.
Additionally, VFP adjusts the image size based on the number of attributes, directly converting all tabular data attributes into images based on their correlations and accounting for their correlation degrees, unlike prior methods with fixed image sizes.

\section{Vortex Feature Positioning} \label{sec: voltex positioning}
As the number of real value attributes in tabular data increases, traditional ML techniques exhibit lower performance and slower training speeds~\cite{borisov2021deep}. CNNs can overcome these limitations by converting tabular data into images and exploiting the benefits of 2-D convolution operations to capture complex patterns in tabular data~\cite{sharma2019deepinsight, zhu2021converting}.

Previous methods of converting tabular data into images aggregate similar features in specific locations. However, because tabular data is inherently heterogeneous, its features do not correspond to pixels in a literal image, where similar pixels are naturally grouped together to form coherent patterns. Aggregating similar features of tabular data in this context may lead to overfitting, especially when dealing with highly correlated features~\cite{yu2003feature, borisov2021deep}. Our new method, Vortex Feature Positioning (VFP), addresses these issues by taking into account two critical factors:
\begin{itemize}
    \item Tabular data is heterogeneous because it comes from distinct sensors, and its features are not like the pixels in an actual image. Therefore, we interpret the functionality of 2-D convolutions as forming appropriate patterns based on tabular features such as numerical values, categories, or textual information rather than merely extracting patterns like an actual image.
    \item Highly correlated features of tabular data should be positioned far away from each other.
\end{itemize}
First, we explain how to embed features into a 2-D matrix while considering convolution operations in Section~\ref{subsec: embed feature}. Then, we describe how to arrange features based on the correlation of attributes in an image in Section~\ref{subsec: arranging features}.

\subsection{Embedding Features Considering Convolution Operations}
\label{subsec: embed feature}
\begin{figure}[ht]
    \centering
    \includegraphics[width=13 cm]{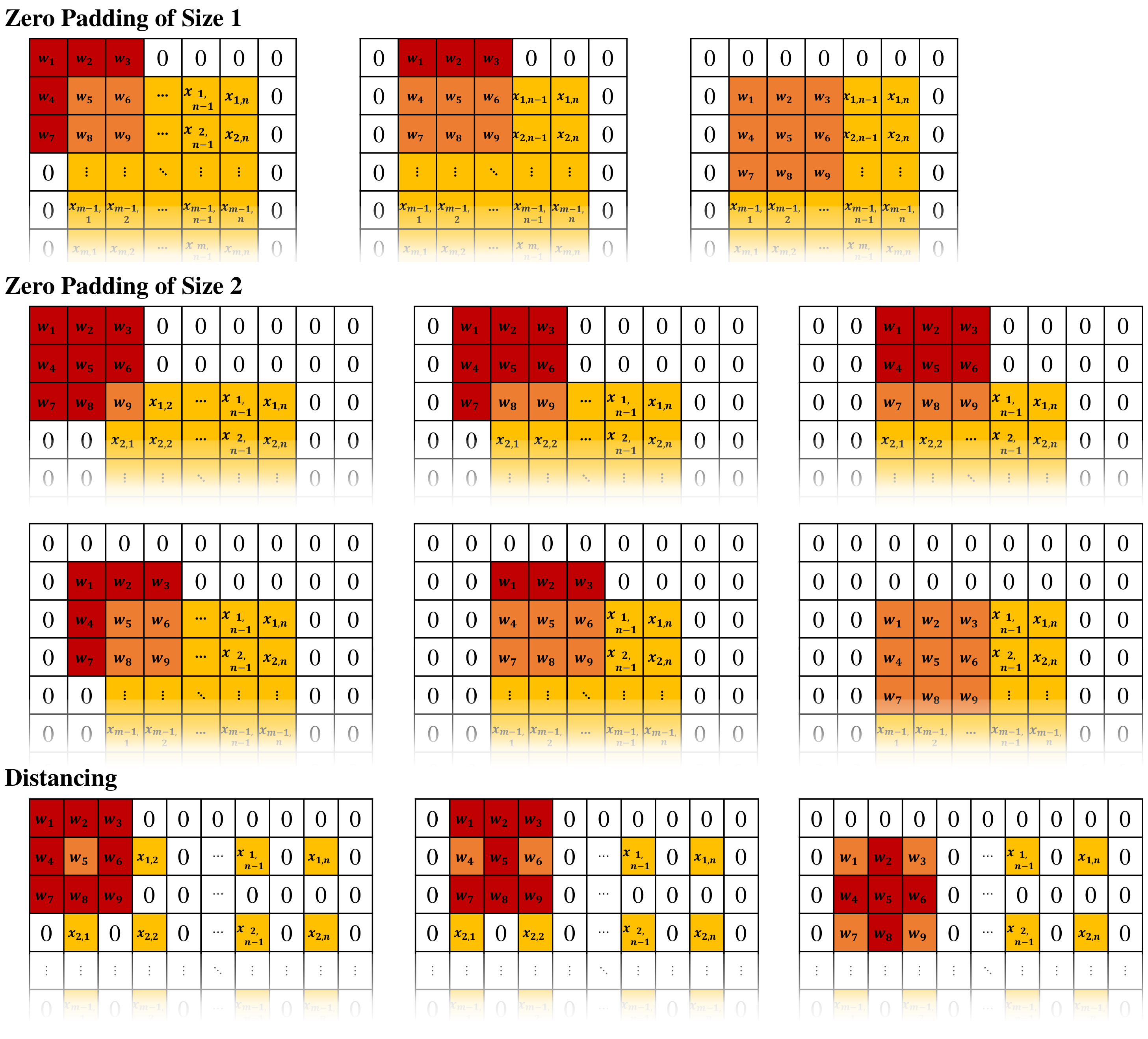}
    \caption{Three cases of feature positioning considering the number of features per convolution operation. From the top: zero padding of sizes 1-2 and distancing.}
    \label{fig:2D Embedding}
\end{figure}

CNNs, such as ResNet~\cite{he2016identity} and DenseNet~\cite{huang2017densely}, typically employ $3\times3$ kernels in the convolution layer. The results of convolution operations with $3\times3$ kernels and feature maps in the first layer impact the kernels of every subsequent layer and, consequently, influence the final inferences~\cite{zeiler2014visualizing}. Therefore, in the first layer, we need to consider the number of features in an image per convolution operation to determine the complexity of patterns. To simplify the explanation, assume there are $m \times n$ attributes in tabular data, where $m$ and $n$ represent the numbers of rows and columns in the feature matrices, respectively. We examine three methods for embedding features into a 2-D matrix: zero-padding with sizes of 1 and 2, as well as distancing, as illustrated in Fig.~\ref{fig:2D Embedding}.

\noindent\textbf{Zero Padding of Size 1 (ZPOS1)}
\begin{equation}
\label{eqn:pad1}
\begin{aligned}
    &{N_{4}^{Pad1}=4} \\            
    &{N_{6}^{Pad1}=2\times{(m-2)}+2\times{(n-2)}=2m+2n-8} \\
    &{N_{9}^{Pad1}=(m-2)\times{(n-2)}=mn-2m-2n+4} 
\end{aligned}
\end{equation}
\textbf{Zero Padding of Size 2 (ZPOS2)}
\begin{equation}
\label{eqn:pad2}
\begin{aligned}
    &{N_{1}^{Pad2}=4}\;\;\;\;\;\;\;\;\;\;{N_{2}^{Pad2}=8}\;\;\;\;\;\;\;\;\;\;{N_{4}^{Pad2}=4} \\ 
    &{N_{3}^{Pad2}=2\times{(m-2)}+2\times(n-2)=2m+2n-8} \\            
    &{N_{6}^{Pad2}=2\times{(m-2)}+2\times(n-2)=2m+2n-8} \\
    &{N_{9}^{Pad2}=(m-2)\times{(n-2)}=mn-2m-2n+4} 
\end{aligned}
\end{equation}
\textbf{Distancing}
\begin{equation}
\label{eqn:distancing}
\begin{aligned}
    &{N_{1}^{Dist}=m\times{n}=mn} \\            
    &{N_{2}^{Dist}=n\times{(m-1)}+m\times{(n-1)}=2mn-m-n} \\
    &{N_{4}^{Dist}=(m-1)\times{(n-1)}=mn-m-n+1}, 
\end{aligned}
\end{equation} 

Here, $N_{i}^{case}$ represents each case's convolution operations involving $i$ features. The $3\times3$ kernels of convolution layers can handle up to nine features. The numbers of convolution operations per the number of features dealt with at once are calculated by Equations~\ref{eqn:pad1}, \ref{eqn:pad2}, and \ref{eqn:distancing}, and the results are shown in Table~\ref{table:convoltion_number}.

\begin{table}[t]
    \small
    \centering 
    \caption{Number of convolution operations per the number of features in cases with zero padding of sizes 1 and 2, and distancing.}
    \label{table:convoltion_number}
    \begin{tabular}{|c|c|c|c|}
    \noalign{\smallskip}\noalign{\smallskip}\hline
    \multirow{2}{*}{\textbf{Cases}} & \textbf{Zero Padding} & \textbf{Zero Padding} & \multirow{2}{*}{\textbf{Distancing}}   \\ 
    \multicolumn{1}{|c|}{\textbf{}}& \textbf{of Size 1}  & \textbf{of Size 2} &   \\ \hline
    \multicolumn{1}{|c|}{\textbf{1 feature}}  & N/A & 4 & $mn$ \\ \hline
    \multicolumn{1}{|c|}{\textbf{2 features}} & N/A & 8 & $2mn-m-n$ \\ \hline
    \multicolumn{1}{|c|}{\textbf{3 features}} & N/A & $2m+2n-8$  & N/A \\ \hline
    \multicolumn{1}{|c|}{\textbf{4 features}} &  4  & 4 & $mn-m-n+1$ \\ \hline            
    \multicolumn{1}{|c|}{\textbf{6 features}} & $2m+2n-8$ & $2m+2n-8$ & N/A \\ \hline
    \multicolumn{1}{|c|}{\textbf{9 features}} & $mn-2m-2n+4$ & $mn-2m-2n+4$ & N/A \\ 
    \Xhline{2\arrayrulewidth}
    \multicolumn{1}{|c|}{\textbf{All cases}} & $mn$ & $mn+2m+2n+4$ & $4mn-2m-2n+1$ \\ \hline
    \end{tabular}
\end{table}

ZPOS1 uses three cases of the number of features for convolution operations totaling $mn$. ZPOS2 uses six cases of features for convolution operations totaling $mn+2m+2n+4$. Distancing uses three cases of the number of features for convolution operations totaling $4mn-2m-2n+1$. The number of total convolution operations denotes how many patterns CNNs have and the number of features used at once, which are related to the complexity of patterns. For example, if nine features ($m=3$ and $n=3$) exist, ZPOS2 and distancing perform twenty-five convolution operations identically. 
However, if there are more than nine features, distancing performs many more convolution operations and uses fewer features than ZPOS2. Therefore, distancing considers many rough patterns between features, while ZPOS2 considers complex patterns as combinations of various features. Since ZPOS1 uses only four, six, and nine features and performs $mn$ convolution operations, it is beneficial for avoiding overfitting compared to ZPOS2, which finds excessively detailed patterns. We also embed features in an image with no padding or distancing. However, convolution operations only use nine features at once, and the number of convolution operations is $(m-2)(n-2)$, the smallest among the methods. Overall, the number of features used in a single convolution operation and the complexity of patterns should be considered when embedding features into a 2-D matrix.

\subsection{Arranging Features Considering Correlations of Attributes} \label{subsec: arranging features}
The traditional ML techniques may exhibit overfitting when heavily relying on highly correlated attributes without dropping them during training~\cite{yu2003feature, grabczewski2005feature}. Previous methods of converting tabular data into images have mainly focused on placing similar features together, resulting in better performance~\cite{sharma2019deepinsight, borisov2021deep}. The high ability of CNNs to capture critical patterns contributes to increased performance. However, we cannot conclude that gathering similar features increases performance because each attribute in tabular data, which is heterogeneous, differs from pixels in an actual image, which represents homogeneous data~\cite{borisov2021deep}.

VFP performs convolution operations on low-correlated attributes by placing features with low correlation at a blank image's center (in ascending order). This is because 2-D convolution operations use center-located features more than edge-located features.

In the context of investigating attribute relationships, the Pearson Correlation Coefficient (PCC), denoted as \( r(\mathbf{a}, \mathbf{b}) \) in Equation \ref{eqn:pearson_coefficient}, is employed. This coefficient measures the linear correlation between two attributes, \( \mathbf{a} \) and \( \mathbf{b} \), which are sets of data points in a dataset. The PCC is calculated by dividing the sum of the products of the deviations of each data point from their respective means (\( \mathbf{a}_i - \overline{\mathbf{a}} \) and \( \mathbf{b}_i - \overline{\mathbf{b}} \)) by the square root of the product of the sums of the squared deviations.
\begin{equation}
\label{eqn:pearson_coefficient}
    r(\mathbf{a},\mathbf{b}) = \frac{\sum\limits^{n}(\mathbf{a}_{i} - \overline{\mathbf{a}})(\mathbf{b}_{i} - \overline{\mathbf{b}})}{\sqrt{
        (\sum\limits^{n}(\mathbf{a}_{i} - \overline{\mathbf{a}})^2)(\sum\limits^{n}(\mathbf{b}_{i} - \overline{\mathbf{b}})^2)
    }},
\end{equation}

Algorithm~\ref{algorithm:reorder} outlines the arranging process, which sorts attributes in ascending or descending order based on \textbf{\textit{the sum of absolute values of PCCs}} aiming to investigate whether gathering similar values improves performance more than spreading them out. 
\begin{algorithm}[t]
\caption{Vortex Feature Positioning: This algorithm arranges attributes based on the sum of Pearson correlation coefficients. It takes a set of attributes $\mathbb{X}$ and outputs the attributes arranged in a vortex pattern $\mathbb{X}_{ordered}$. The vector $\mathbf{c}$ contains the sum of absolute Pearson correlation coefficients for each feature $c_i$, which determines the order.}
\begin{algorithmic}[1]
    \STATE \textbf{Input:} $\mathbb{X}$ = \{$\mathbf{x}_{i}|i=1,2...,k$\}
    \STATE \textbf{Output:} $\mathbb{X}_{ordered}$ \\              
    \STATE $\mathbf{c} = $\{$c_i|i=1,2,...,k$\} \\
     \FOR {$i \in 1,2,...,k$} {
    \STATE    $c_{i} \gets 0$ 
    \FOR {$j \in 1,2,...,k$} {
    \STATE        $c_{i} \gets c_{i}+|r(\mathbf{x}_i,\mathbf{x}_j)|$
        }     
    \ENDFOR
    }
    \ENDFOR 
    \STATE ${\mathbb{X}_{ordered}} \gets \mathbb{X}[rank_{ascending}(\mathbf{c})]$ or $\mathbb{X}[rank_{descending}(\mathbf{c})]$ 
\end{algorithmic}
\label{algorithm:reorder}
\end{algorithm}

$\mathbb{X}$ is a tabular dataset composed of attributes $\mathbf{x}_{i}$ (\textit{line 1} in Algorithm~\ref{algorithm:reorder}). $\mathbf{c}$ is a vector composed of $c_i$, which is the sum of PCCs of $\mathbf{x}_{i}$ and other attributes (\textit{line 3}). $rank_{ascending}(\mathbf{c})$ and $rank_{descending}(\mathbf{c})$ are functions that arrange $\mathbf{c}$ according to $c_i$ size in ascending or descending order and provide ordered indices (\textit{lines 4-9}). We arrange $\mathbb{X}$ based on the order stored in $\mathbf{c}$ (\textit{line 10}).

After arranging attributes based on the PCC, features of attributes are embedded in an image. CNNs perform more convolution operations on features at the center of the image than on exterior features.
Therefore, placing features in a vortex shape from the center of the image exploits desired features for many convolution operations, while undesired features are the opposite.
For example, arranging features in ascending order according to the sum of PCCs performs many convolution operations with low-correlated features such as $x_{1}^{c}$, $x_{2}^{c}$, $x_{3}^{c}$, and $x_{4}^{c}$, which are similar to feature selection in ML techniques for tabular data~\cite{grabczewski2005feature}. In contrast, features $x_{n}^{c}$ and $x_{n-1}^{c}$, which have high PCCs and are located near the edge, are used for relatively fewer convolution operations.

\begin{figure}[ht]
    \centering
    \includegraphics[width=\linewidth]{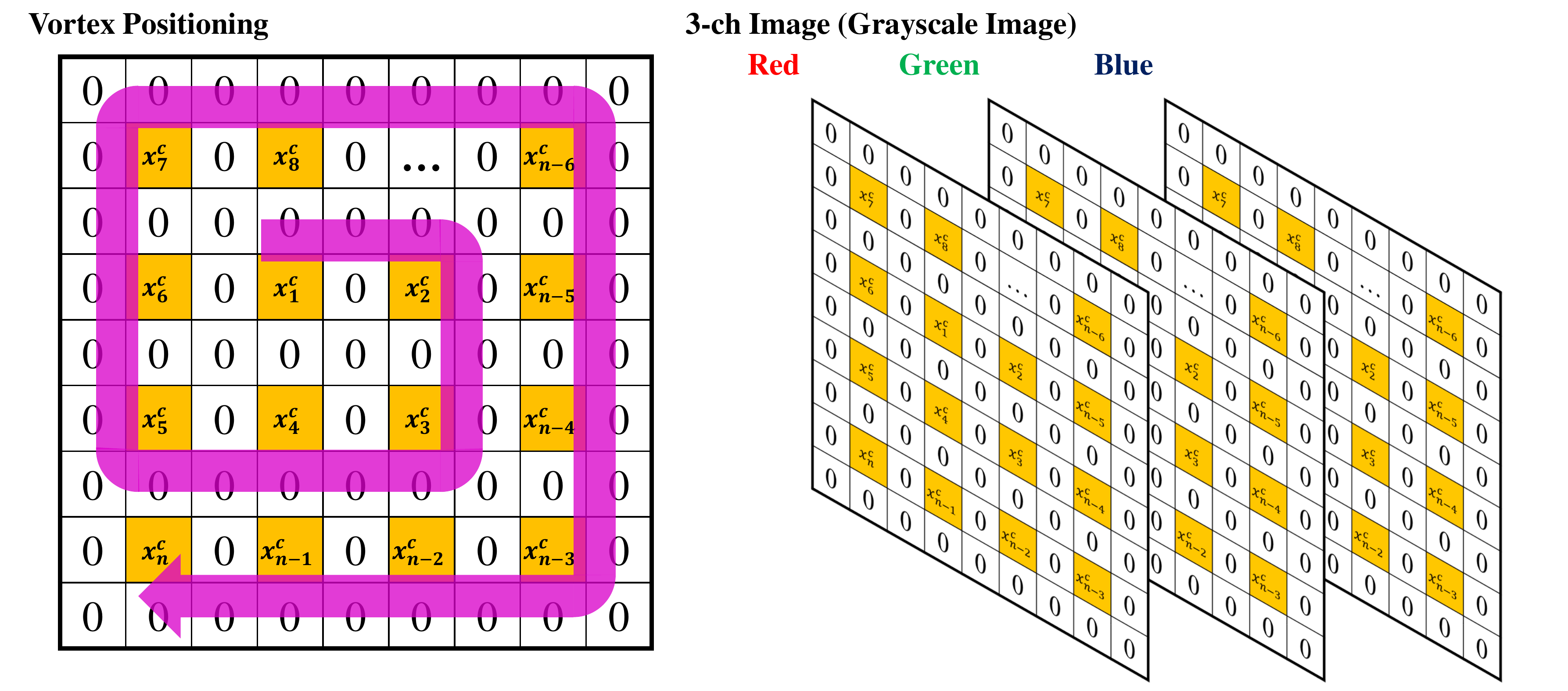}
    \caption{Vortex Feature Positioning and forming a 3-channel image by copying a 2-D matrix.}
    \label{fig:channel Positioning}
\end{figure}

We use a single-channel image (i.e., a grayscale image) in VFP. However, state-of-the-art CNNs (i.e., ResNet and DenseNet) require 3-channel images. This paper employs Pre-activated ResNet-18~\cite{he2016identity}. Therefore (R), green (G), and blue (B) channels all have the same feature map, as shown in Fig.~\ref{fig:channel Positioning}.

\section{Analysis of Vortex Feature Positioning} \label{sec: convergence}
\subsection{Preliminaries: Vortex Feature Positioning}
Vortex Feature Positioning (VFP) converts tabular data into an image format by positioning attributes outward from the center. On average, features closer to the center are less correlated with others, creating a spatial representation of feature correlations.

\subsection{Assumptions}
For the subsequent analysis, we'll state our assumptions clearly: 
\begin{itemize}
    \item \textbf{Locally Convex Regions}: While the entire loss surface of a CNN trained on VFP images might be non-convex, regions close to the center (where less correlated attributes reside) are approximated as locally convex.
    \item \textbf{Bounded Gradients}: The gradient magnitude does not grow indefinitely. This is a reasonable assumption given the structured nature of VFP images and the bounded activations in CNNs.
    \item \textbf{Lipschitz Continuity of Gradients}: Despite the spatial rearrangements due to VFP, we assume that small changes in the input lead to proportionally small changes in the output. This is represented by a Lipschitz constant $L$ such that:
    \[
    \|\nabla f(x) - \nabla f(y)\| \leq L \|x - y\|
    \]
    \item \textbf{Diagonal Dominance in Hessian for Central Features}: For features positioned centrally in VFP images (and thus less correlated), their Hessian matrix is approximately diagonal, indicating limited interactions with other features.
    Consider two features, $ x_i $ and $ x_j $, with their effect on the loss $ f $ represented as a second-order term in the Hessian matrix $ H $:
    \[
    H_{ij} = \frac{\partial^2 f}{\partial x_i \partial x_j}
    \]
    For uncorrelated features $x_i$ and $x_j$, $H_{ij}$ is close to zero. A function is convex if its Hessian is positive semi-definite everywhere. With VFP, central features (less correlated) have a Hessian structure that's approximately diagonal:
    \[
    \mathbf{H} \approx \text{diag}(\lambda_1, \lambda_2, ..., \lambda_k), \quad \lambda_i \geq 0
    \]
    where $\lambda_i$ are eigenvalues. Local convexity around these features is suggested if these eigenvalues are non-negative.
\end{itemize}

\subsection{Convolution's Role in Convergence}
In CNNs, convolution operations, especially in the initial layers, focus on the local regions of the input image. Given the VFP structure, these convolutions will predominantly operate on the less correlated central features in their early stages. This behavior ensures:
\begin{itemize}
    \item \textbf{Unique Feature Capture}: Convolutions can effectively capture the unique characteristics of these less correlated features, providing more informative gradient signals during backpropagation.
    \item \textbf{Gradient Quality}: Gradients derived from less correlated features are likely more distinct, reducing the chances of vanishing or exploding gradients, especially in the early layers.
\end{itemize}

\subsection{Correlation-Induced Convexity}
The combined influence of less correlated features $x_i$ and $x_j$ on the loss $f$ can be depicted by the Hessian entry $H_{ij}$, which is expected to be close to zero. The diagonal-like Hessian for central features in VFP images suggests potential local convex regions.

\subsection{SGD Convergence Analysis with VFP Images}
In the context of VFP, we analyze the convergence of Stochastic Gradient Descent (SGD) by incorporating the spatial structure of VFP images into the gradient computation.

Consider the general SGD update rule:
\[
\theta_{t+1} = \theta_t - \eta_t \nabla f(x_t, \theta_t),
\]
where \(\theta_t\) are the model parameters, and \(x_t\) is the feature vector at iteration \(t\), \(\eta_t\) is the learning rate, and \(\nabla f(x_t, \theta_t)\) is the gradient of the loss function with respect to both \(x\) and \(\theta\) at iteration \(t\). In the case of VFP, the gradient is modified to:
\[
\nabla f(x_i, \theta_t) = \frac{\partial f}{\partial x_i}(\theta_t) + \sum_{j \neq i} C_{ij} \frac{\partial f}{\partial x_j}(\theta_t),
\]
where \(C_{ij}\) represents the interaction coefficient between features \(x_i\) and \(x_j\) based on their spatial positioning in the VFP image. The SGD update rule, considering the VFP structure, is thus:
\[
\theta_{t+1} = \theta_t - \eta_t \left( \frac{\partial f}{\partial x_t}(\theta_t) + \sum_{j \neq t} C_{tj} \frac{\partial f}{\partial x_j}(\theta_t) \right).
\]

To derive the convergence rate, we start by considering the squared norm of the parameter update difference:
\[
\|\theta_{t+1} - \theta^*\|^2 = \|\theta_t - \eta_t \nabla f(x_t, \theta_t) - \theta^*\|^2,
\]
which expands to:
\[
\|\theta_{t+1} - \theta^*\|^2 = \|\theta_t - \theta^*\|^2 - 2\eta_t \nabla f(x_t, \theta_t)^\top (\theta_t - \theta^*) + \eta_t^2 \|\nabla f(x_t, \theta_t)\|^2.
\]
Applying the convexity inequality, we get:
\[
f(x^*, \theta^*) \geq f(x_t, \theta_t) + \nabla f(x_t, \theta_t)^\top (\theta^* - \theta_t),
\]
leading to:
\[
\nabla f(x_t, \theta_t)^\top (\theta_t - \theta^*) \geq f(x_t, \theta_t) - f(x^*, \theta^*).
\]
Substituting into the squared norm equation and assuming a bound on the gradient norm \( \|\nabla f(x_t, \theta_t)\|^2 \leq G^2 \), we obtain:
\[
\|\theta_{t+1} - \theta^*\|^2 \leq \|\theta_t - \theta^*\|^2 - 2\eta_t (f(x_t, \theta_t) - f(x^*, \theta^*)) + \eta_t^2 G^2.
\]
Rearranging and dividing by \(2\eta_t\), the expected sub-optimality is given by:
\[
f(x_t, \theta_t) - f(x^*, \theta^*) \leq \frac{1}{2\eta_t}(\|\theta_t - \theta^*\|^2 - \|\theta_{t+1} - \theta^*\|^2) + \frac{\eta_t}{2} G^2.
\]
Summing this inequality over \(t\) from \(1\) to \(T\), we get:
\[
\sum_{t=1}^{T} (f(x_t, \theta_t) - f(x^*, \theta^*)) \leq \frac{1}{2\eta_t} \sum_{t=1}^{T} (\|\theta_t - \theta^*\|^2 - \|\theta_{t+1} - \theta^*\|^2) + \frac{\eta_t G^2 T}{2}.
\]
The summation of the squared norm differences forms a telescoping series, simplifying to:
\[
\sum_{t=1}^{T} (\|\theta_t - \theta^*\|^2 - \|\theta_{t+1} - \theta^*\|^2) = \|\theta_1 - \theta^*\|^2 - \|\theta_{T+1} - \theta^*\|^2.
\]
Assuming \( \|\theta_1 - \theta^*\|^2 \) is bounded and \( \|\theta_{T+1} - \theta^*\|^2 \geq 0 \), the summation is bounded above by \( \|\theta_1 - \theta^*\|^2 \). Therefore:
\[
\sum_{t=1}^{T} (f(x_t, \theta_t) - f(x^*, \theta^*)) \leq \frac{\|\theta_1 - \theta^*\|^2}{2\eta_t} + \frac{\eta_t G^2 T}{2}.
\]
Dividing by \(T\) to obtain the average expected sub-optimality over \(T\) iterations, we get:
\[
\frac{1}{T}\sum_{t=1}^{T} \mathbb{E}[f(x_t, \theta_t) - f(x^*, \theta^*)] \leq \frac{\|\theta_1 - \theta^*\|^2}{2\eta_t T} + \frac{\eta_t}{2} G^2.
\]
Assuming a diminishing learning rate, typically

 \(\eta_t = \frac{\eta}{\sqrt{t}}\), the bound becomes:
\[
\mathbb{E}[f(x_t, \theta_t) - f(x^*, \theta^*)] \leq \frac{\|\theta_1 - \theta^*\|^2}{2\eta \sqrt{t}} + \frac{\eta G^2}{2\sqrt{t}}.
\]
As \(t\) grows, the term \(\frac{1}{\sqrt{t}}\) dominates, leading to the convergence rate of:
\[
\mathbb{E}[f(x_t, \theta_t) - f(x^*, \theta^*)] = O\left(\frac{1}{\sqrt{t}}\right).
\]
This bound indicates that the expected sub-optimality of SGD with VFP images converges at a rate of \(O\left(\frac{1}{\sqrt{t}}\right)\), typical for SGD in convex optimization scenarios with a diminishing learning rate.

\subsection{Summary}
Through VFP, tabular data is transformed into images with a spatial structure that enables convolution operations in CNNs to capture unique feature characteristics effectively. The combined advantages of VFP and the inherent properties of CNNs, along with informed assumptions, suggest that the convergence properties of SGD remain theoretically robust in this setup. This hybrid approach could improve optimization strategies for non-traditional complex IIoT data.

\section{Experimental Evaluation} \label{sec: evaluation}
In this evaluation, we analyze seven benchmark tabular datasets, categorized by their number of attributes and samples, sourced from the UCI Machine Learning Repository~\cite{uci_ml_repo} and Kaggle~\cite{kaggle}. These datasets include Iris, Wine, Dry Bean, and SECOM, which are widely utilized in industrial applications; the Epileptic Seizure dataset, known for its challenges in medical data training; the HELOC dataset, which illustrates the disparity between the current market value and the purchase price of homes; and the HIGGS dataset, used to differentiate between processes that produce Higgs bosons and those that do not.

Each dataset's information is summarized in Table~\ref{table:data description}. Iris, Wine, Dry Bean, HELOC, and HIGGS have relatively small numbers of attributes, while Epileptic Seizure and SECOM have 178 and 591 attributes, respectively. All datasets contain real-valued attributes, which are challenging to train using ML techniques for tabular data. Each dataset is randomly split into training and testing sets in a 0.8:0.2 ratio, and a Min-Max scaler is applied to normalize the data.

\begin{figure*}[t]
    \centering
    \includegraphics[width=\linewidth]{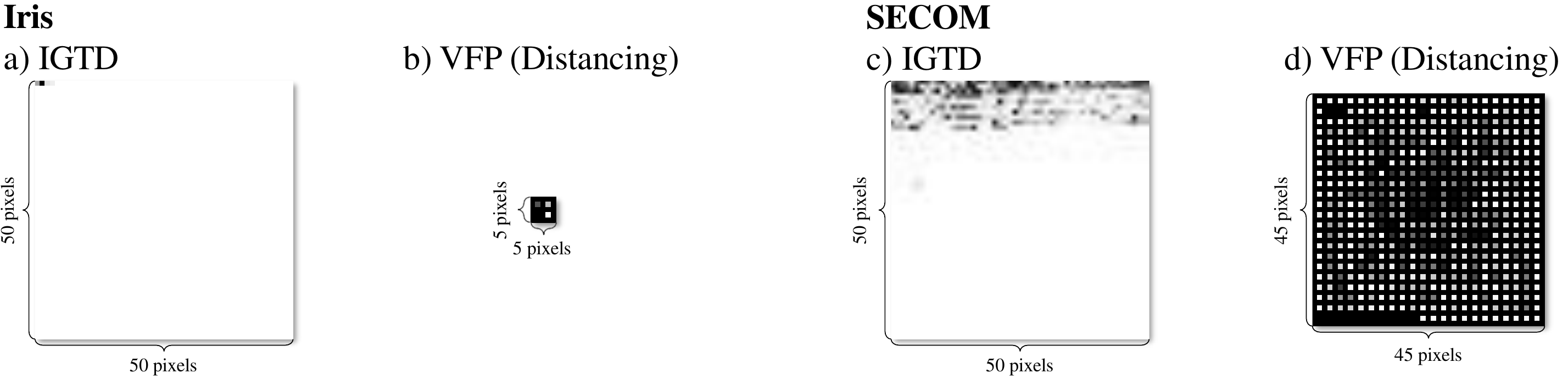}
    \caption{Converted images of IGTD and VFP (with distancing) using Iris and SECOM datasets.}
    \label{fig:image data}
\end{figure*}

\begin{table}[t]
    \centering 
    \caption{Testing accuracies (with standard deviations) using VFP. Feature arrangement methods include Ascending (less correlated features near the center) and descending (more correlated features near the center). Each case was tested with eight trials, and the cases achieving the top two accuracies for each dataset are bolded.}
    \label{table:test_result_vfp_only}
    \includegraphics[width=\linewidth]{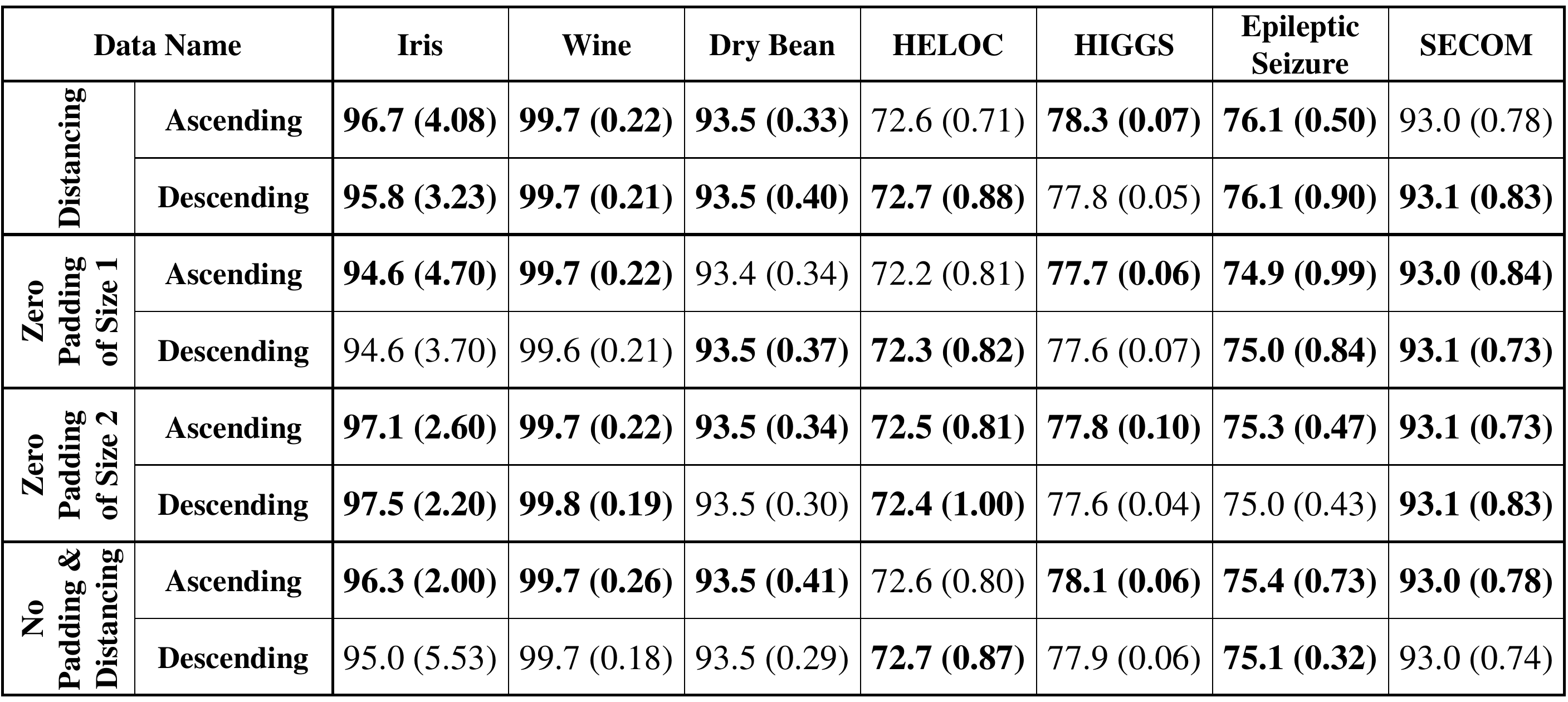}
\end{table}
\begin{table}[t]
    \centering 
    \caption{Testing accuracies (with standard deviations) of VFP, traditional machine learning techniques, and previous deep learning-based image conversion techniques. The top three results for each dataset are bolded.}
    \label{table:test_result_vfp_ml_others}
    \includegraphics[width=\linewidth]{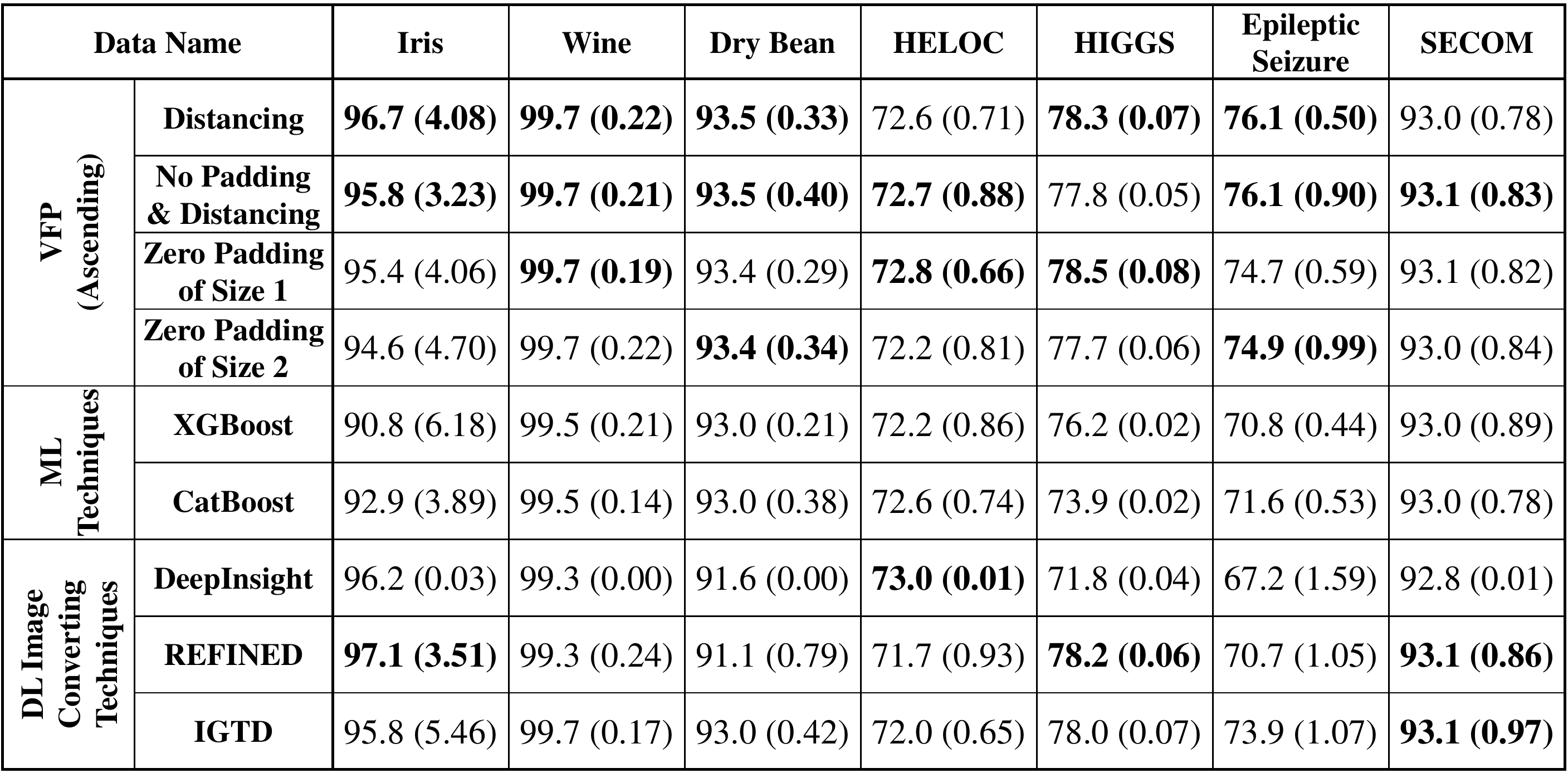}
\end{table}

\begin{figure}[ht]
    \centering
    \includegraphics[width=\linewidth]{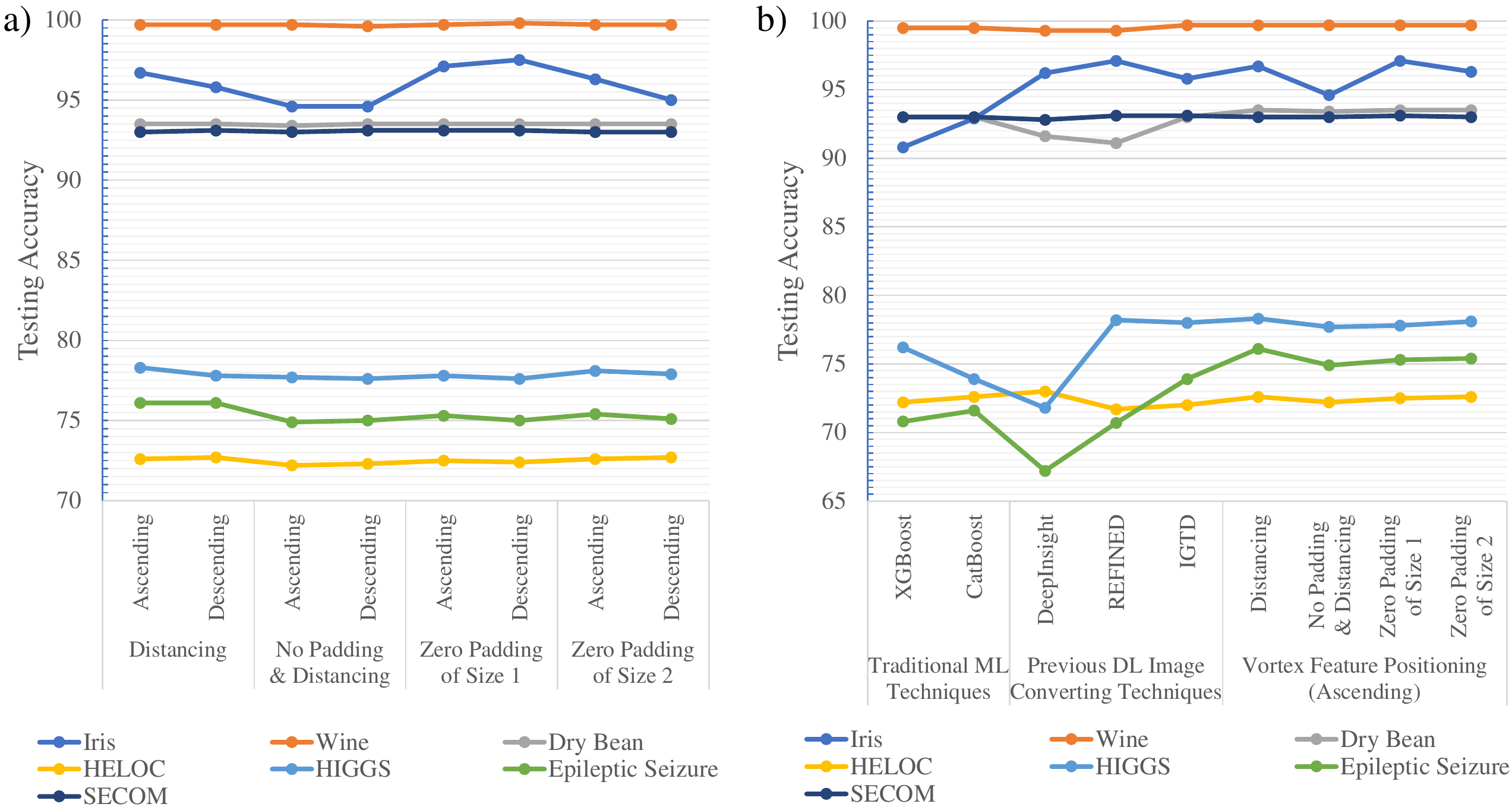}
    \caption{Part \textbf{a)} lines correspond to the test accuracies presented in Table~\ref{table:test_result_vfp_only}, while part \textbf{b)} lines relate to the test accuracies in Table~\ref{table:test_result_vfp_ml_others}. VFP consistently shows better testing performance than other methods.}
    \label{fig:result graph}
\end{figure}

Fig.~\ref{fig:image data} shows converted images for the Iris and SECOM datasets using IGTD and VFP with distancing. IGTD~\cite{zhu2021converting} is a recent method for converting tabular data into images that has shown better performance and smaller image sizes than other methods such as DeepInsight~\cite{sharma2019deepinsight} and REFINED~\cite{bazgir2020representation}. The image size generated by IGTD is fixed at $50 \times 50$ pixels, regardless of the number of attributes in the dataset. Although the authors~\cite{zhu2021converting} mention that the image size can be changed according to the number of attributes, there is no proposed generalized rule for image sizes across different datasets. In contrast, VFP, with distancing, adjusts the image size based on the number of attributes, resulting in image sizes of $5 \times 5$ and $45 \times 45$ pixels for the Iris and SECOM datasets, respectively. 

\subsection{Models} \label{subsec: model}
We evaluate the performance using Pre-activated ResNet-18, constructed by stacking residual blocks. He et al. used four types of residual blocks~\cite{he2016identity} to construct ResNets. We use the simplest ResNet variant to show the effectiveness of CNNs compared to ML techniques for tabular data. One modification in our model is that the size of kernels in the first layer is changed to $3 \times 3$ with a stride of 1 to be consistent with VFP's convolution operations. 

\subsection{Experimental Setups} \label{subsec: evaluation}
Our experiments were conducted on an NVIDIA A100 40GB GPU. We employed a cosine annealing warm-up with restarts for optimization, setting the learning rate to vary between 0.01 and 0.001. This learning rate is reset every five epochs. The experiments were run with a mini-batch size of 128, and each dataset was trained for 200 epochs, except for the HIGGS dataset, which was trained for 20 epochs. To ensure the objectivity of our testing results, each experiment was repeated eight times with different random seeds, ranging from 1000 to 8000 in 1000 increments.

\subsubsection{Converting Tabular Data into Images}
To convert tabular data into images, we first embed one sample of each tabular dataset into a 2-D matrix with different types of zero padding and distancing: zero padding of size 1 (ZPOS1), zero padding of size 2 (ZPOS2), and distancing. We then convert the resulting 2-D matrix into a 3-channel image by copying itself.

\subsubsection{Employing Vortex Feature Positioning}
VFP improves performance by conducting convolution operations on selected features. The test results, shown in Table~\ref{table:test_result_vfp_only} and in part \textbf{a)} of Fig.~\ref{fig:result graph}, demonstrate that arranging features in ascending order based on their correlation from the center leads to better performance. Additionally, the choice of strategy—whether employing distancing, zero padding of sizes 1 and 2, or opting for no padding—varies depending on the datasets' complexity.

Significantly, as emphasized in Table~\ref{table:test_result_vfp_only}, which showcases the top two testing accuracies for each scenario (bolded and categorized by dataset and positioning method), distancing consistently delivers moderately good results. This is the case even though distancing uses only 1, 2, or 4 features, which results in simpler patterns. This simplicity is beneficial as it helps avoid overfitting, thereby boosting overall performance.

Consequently, a tailored embedded space capitalizing on specific feature relationships will produce more effective patterns and enhance performance.

\subsubsection{Comparison Performance with Machine Learning Techniques for Tabular Data}
We evaluated the effectiveness of VFP against traditional machine learning techniques (XGBoost and CatBoost) as well as previously established deep learning conversion methods (DeepInsight, REFINED, and IGTD). For traditional machine learning techniques, we modified the maximum number of leaves in CatBoost from 16 to 64 and adjusted the minimum child weight in XGBoost from 1 to 9. Additionally, we varied the maximum tree depth from 4 to 16 for both CatBoost and XGBoost. For established deep learning conversion methods, we applied the same training configuration as that used in VFP cases.

These comparisons are detailed in Table~\ref{table:test_result_vfp_only} and illustrated in part \textbf{b)} of Fig.~\ref{fig:result graph}. For the machine learning techniques, we conducted eight repeated experiments at various tree depths, ranging from one to sixteen. Similarly, we repeated the tests eight times for the deep learning approaches, maintaining the same training conditions used for VFP.

The results indicate that VFP surpasses other methods across most of the datasets tested. Particularly, VFP achieved test accuracy improvements ranging from 2.2\% to 10.9\% higher than other methods on the Epileptic Seizure dataset. This indicates that VFP is the most effective among the known conversion methods and excels in optimizing image sizes for training, thereby minimizing computational resource wastage.

Overall, our evaluation demonstrates the robustness and superiority of VFP in handling diverse datasets, offering a significant improvement over traditional machine learning techniques and existing deep learning-based image conversion methods. The flexibility in adjusting image sizes based on the number of attributes and the strategic arrangement of features contributes to VFP's enhanced performance.

By capitalizing on the strengths of both convolutional neural networks and VFP's unique feature arrangement strategy, our method provides a reliable and efficient approach to transforming tabular data into a format that maximizes the potential of deep learning models. This leads to better generalization, improved accuracy, and a reduction in overfitting, making VFP a valuable tool for a wide range of applications in industrial and other domains.

\section{Conclusions} \label{sec: conclusion}
This paper introduces Vortex Feature Positioning (VFP), a method for converting tabular data into images based on attribute correlations. By utilizing ascending orders and strategic positioning, VFP achieves superior performance compared to traditional ML techniques across various datasets.

Compared to existing methods for converting tabular data into images, VFP is optimized to avoid overfitting by considering the correlations among attributes in the dataset. Additionally, VFP can adjust the size of the converted images to accommodate different numbers of attributes, providing a flexible and adaptable solution for industrial tabular data, including IIoT data.

VFP's ability to transform tabular data into a format that leverages the strengths of convolutional neural networks significantly enhances model performance. Our experiments demonstrate that VFP consistently outperforms both traditional machine learning methods and other deep learning-based image conversion techniques. This makes VFP a valuable tool for a wide range of applications in industrial and other domains, where handling complex, high-dimensional tabular data is crucial.

Overall, VFP represents a robust and efficient approach for enhancing the analysis of tabular data, paving the way for more effective and scalable machine-learning solutions in diverse fields. Future work will explore further optimization techniques and applications to expand the utility and impact of VFP in various data-driven industries.

\bibliographystyle{elsarticle-num-names}
\bibliography{Mybib}

\end{document}